\documentclass{article}

\usepackage{arxiv}

\usepackage[dvipsnames]{xcolor}
\usepackage[utf8]{inputenc} 
\usepackage[T1]{fontenc}    
\usepackage{hyperref}       
\usepackage{url}            
\usepackage{booktabs}       
\usepackage{amsfonts}       
\usepackage{nicefrac}       
\usepackage{microtype}      
\usepackage{lipsum}		
\usepackage{graphicx}
\usepackage[authoryear]{natbib}
\usepackage{tikz}
\usepackage{amsmath}
\usepackage{algorithm}
\usepackage{algpseudocode}
\usepackage{wrapfig}

\hypersetup{
	colorlinks=true,
	linkcolor=blue!60!Cerulean!90!black,
	citecolor=blue!60!Cerulean!90!black,
	urlcolor=magenta,
	pdftitle={Qgraph-bounded Q-learning: Stabilizing Model-Free Off-Policy Deep Reinforcement Learning},
}

\definecolor{sns1}{rgb}{0.12156862745098039, 0.4666666666666667, 0.7058823529411765}
\definecolor{sns2}{rgb}{1.0, 0.4980392156862745, 0.054901960784313725}
\definecolor{sns3}{rgb}{0.17254901960784313, 0.6274509803921569, 0.17254901960784313}
\definecolor{sns4}{rgb}{0.8392156862745098, 0.15294117647058825, 0.1568627450980392}
\definecolor{sns5}{rgb}{0.5803921568627451, 0.403921568627451, 0.7411764705882353}
\definecolor{sns6}{rgb}{0.5490196078431373, 0.33725490196078434, 0.29411764705882354}

\title{Qgraph-bounded Q-learning: Stabilizing Model-Free Off-Policy Deep Reinforcement Learning}

\date{July 2020}	

\author{Sabrina Hoppe
	\\
	Corporate Research\\
	Robert Bosch  GmbH\\
	71272 Renningen\\
	\texttt{sabrina.hoppe@de.bosch.com} \\
	\And
	Marc Toussaint\\
	Learning and Intelligent Systems Lab\\
	TU Berlin\\
	10587 Berlin\\
	\texttt{toussaint@tu-berlin.de}
}

\renewcommand{\headeright}{}
\renewcommand{\undertitle}{Technical Report}

\hypersetup{
pdftitle={Qgraph-bounded DDPG},
pdfauthor={Sabrina Hoppe, Marc Toussaint},
pdfkeywords={Deep Reinforcement Learning, DDPG},
}

\newcommand{\qgraph}{Q\textsc{graph}}
\newcommand{\ourmethod}{\qgraph-bounded Q-learning}

\def\equationautorefname~#1\null{%
	Eq.~(#1)\null
}

\begin{document}
\maketitle

\begin{abstract}
	In state of the art model-free off-policy deep reinforcement learning, a replay memory is used to store past experience and derive all network updates.
Even if both state and action spaces are continuous, the replay memory only holds a finite number of transitions. 
We represent these transitions in a \textit{data graph} and link its structure to soft divergence.
By selecting a subgraph with a favorable structure, we construct a simplified Markov Decision Process for which exact Q-values can be computed efficiently as more data comes in.
The subgraph and its associated Q-values can be represented as a \qgraph.
We show that the Q-value for each transition in the simplified MDP is a lower bound of the Q-value for the same transition in the original continuous Q-learning problem.
By using these lower bounds in temporal difference learning, our method QG-DDPG is less prone to soft divergence and exhibits increased sample efficiency while being more robust to hyperparameters.
\qgraph{}s also retain information from transitions that have already been overwritten in the replay memory, which can decrease the algorithm's sensitivity to the replay memory capacity.

\end{abstract}


\section{Introduction}
With the wide-spread success of neural networks, also deep reinforcement learning (RL) has enabled rapid improvements in many domains including computer games~\citep{silver2017mastering} and simulated continuous control tasks~\citep{mnih2016asynchronous}.
Particularly in areas where correct environment models are hard to obtain, such as robotic manipulation, model-free approaches have the potential to outperform model-based solutions \citep{fazeli2017learning,Levine2016LearningHC} -- as long as enough training data is available or can be generated.

From a theoretical point of view, deep reinforcement learning is still under-investigated, in particular deep Q-learning and DDPG.
While Q-learning is known to have convergence issues even with linear function approximation~\citep{baird1995residual}, deep Q-learning combines highly non-linear function approximation with off-policy learning and bootstrapping -- a combination that has been termed \textit{deadly triad} by~\citet{sutton2018reinforcement} because of the instabilities it is likely to induce.
Empirically, deep Q-learning does not seem to fully exhibit these expected divergence issues~\citep{van2018deep} but its performance can be unreliable and hard to reproduce~\citep{henderson2018deep}.

The contribution in this work is two-fold:\\
To add to the community's understanding of when deep Q-learning diverges, we first propose a graph-perspective on the replay memory (\textit{data graph}) which allows to analyze its structure and show on an educational example that specific types of structures are linked to divergence.\\
Second, we introduce a \qgraph: a subgraph that was chosen such that exact Q-values for the induced finite Markov Decision Process (MDP) can be computed using Q-iteration. 
We show that these Q-values are lower bounds for the Q-values in the original MDP that models a continuous learning problem.
Using these bounds in temporal difference learning stabilizes deep reinforcement learning for continuous state and action spaces through DDPG by preventing cases of divergence.
Further analyses reveal that this increases sample efficiency, robustness to hyperparameters and preserves information from transitions that have already been overwritten in the replay memory.

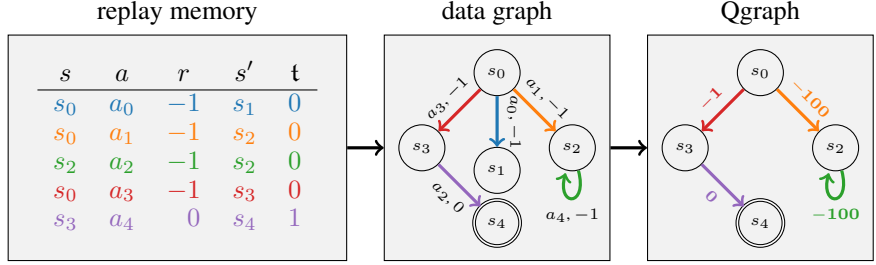
\begin{figure*}
	\centering
	\begin{tikzpicture}
	%
	\def\blockgap{0.5}
	\node(replaymemtitle) at (-6, 1.8) {replay memory};
	\node[rectangle, draw, minimum height=3cm, minimum width=4.5cm, fill=black!5](replaymem) at (-6, 0) {\begin{tabular}{ccccc}
		$s$ & $a$ & $r$ & $s'$ & $\mathfrak{t}$ \\
		\hline
		\textcolor{sns1}{$s_0$} & \textcolor{sns1}{$a_0$} &\textcolor{sns1}{$-1$} & \textcolor{sns1}{$s_1$} & \textcolor{sns1}{$0$} \\
		\textcolor{sns2}{$s_0$} & \textcolor{sns2}{$a_1$} &\textcolor{sns2}{$-1$} & \textcolor{sns2}{$s_2$} & \textcolor{sns2}{$0$} \\
		\textcolor{sns3}{$s_2$} & \textcolor{sns3}{$a_2$} &\textcolor{sns3}{$-1$} & \textcolor{sns3}{$s_2$} & \textcolor{sns3}{$0$} \\
		\textcolor{sns4}{$s_0$} & \textcolor{sns4}{$a_3$} &\textcolor{sns4}{$-1$} & \textcolor{sns4}{$s_3$} & \textcolor{sns4}{$0$} \\
		\textcolor{sns5}{$s_3$} & \textcolor{sns5}{$a_4$} &\textcolor{sns5}{$~~~0$} & \textcolor{sns5}{$s_4$} & \textcolor{sns5}{$1$} \\
		\end{tabular}};
	%
	\def\xmid{-6+2.25+1.5+\blockgap}
	\node(datagraphtitle) at (\xmid, 1.8) {data graph};
	\node[rectangle, draw, minimum height=3cm, minimum width=3cm, fill=black!5](datagraph) at (\xmid, 0) {};
		\node[circle, draw] (1) at (\xmid, 1) {\tiny $s_0$};
		\node[circle, draw] (2) at (\xmid, -0.3) {\tiny $s_1$};
		\node[circle, draw] (3) at (\xmid + 1, 0) {\tiny $s_2$};
		\node[circle, draw] (4) at (\xmid - 1, 0) {\tiny $s_3$};
		\node[circle, draw, double] (5) at (\xmid, -1) {\tiny $s_4$};
		\path (1) edge[->, draw=sns1, very thick] node[sloped, above] {\tiny $a_0$, $-1$} (2)
			  (1) edge[->, draw=sns2, very thick] node[sloped, above] {\tiny $a_1$, $-1$} (3)
			  (1) edge[->, draw=sns4, very thick] node[sloped, above] {\tiny $a_3$, $-1$} (4)
			  (4) edge[->, draw=sns5, very thick] node[sloped, below] {\tiny $a_2$, $0$} (5)
			  (3) edge[->, draw=sns3, very thick, loop below] node[auto] {\tiny $a_4$, $-1$} (3)
		;
	%
	\def\qmid{\xmid + 1.5 + \blockgap + 1.5}
	\node(qgraphtitle) at (\qmid, 1.8) {Qgraph};
	\node[rectangle, draw, minimum height=3cm, minimum width=3cm, fill=black!5](qgraph) at (\qmid, 0) {};
	\node[circle, draw] (q1) at (\qmid, 1) {\tiny $s_0$};
	\node[circle, draw] (q3) at (\qmid + 1, 0) {\tiny $s_2$};
	\node[circle, draw] (q4) at (\qmid - 1, 0) {\tiny $s_3$};
	\node[circle, draw, double] (q5) at (\qmid, -1) {\tiny $s_4$};
	\path (q1) edge[->, draw=sns2, very thick] node[sloped, above] {\textcolor{sns2}{\tiny $\mathbf{-100}$}} (q3)
	(q1) edge[->, draw=sns4, very thick] node[sloped, above] {\textcolor{sns4}{\tiny $\mathbf{-1}$}} (q4)
	(q4) edge[->, draw=sns5, very thick] node[sloped, below] {\textcolor{sns5}{\tiny $\mathbf{0}$}} (q5)
	(q3) edge[->, draw=sns3, very thick, loop below] node[auto] {\textcolor{sns3}{\tiny $\mathbf{-100}$}} (q3)
	;
	%
	%
	\path (replaymem) edge[->, very thick] (datagraph)
	(datagraph) edge[->, very thick] (qgraph)
	;
	\end{tikzpicture}
	\caption{We represent the replay memory (left) as a data graph (middle) and extract a subgraph (right) such that its structure allows to compute exact Q-values using Q-iteration for the resulting finite MDP.
	}
	\label{fig:teaser}
\end{figure*}

\section{Preliminaries}
We consider a standard reinforcement learning setup where an agent interacts in discrete time steps $t=1, \dots, T$ with an environment that is modeled as a Markov Decision Process (MDP) with state space $\mathcal{S}$, action space $\mathcal{A}$, initial state distribution $p_0(s)$, transition dynamics $p(s_{t+1} | s_t, a_t)$ and a reward function $r(s_t, a_t)$.
In the following, we will assume deterministic transition dynamics; but the empirical evaluation will come back to the case of non-deterministic transitions.

At each time step $t$, the agent can observe its state $s_t$ and take an action $a_t$ which determines the next state $s_{t+1}$ and an associated reward $r_t$.
A policy is a function $\pi$ that maps from states to actions.
The sum over future expected rewards when following policy $\pi$ starting from state $s_t$ is called return: $R^\pi_t = \sum_t^\infty \gamma^t r_i$, where $\gamma$ is the so-called discount factor.
For $\gamma < 1$ and a constant reward $r$ on infinite trajectories, the return forms a geometric series and converges to $\frac{r}{1-\gamma}$.
Thus, if the reward function is bounded by $r_\text{min}$ and $r_\text{max}$, the range of possible Q-value can be bounded as follows~\citep{lee2015tighter}:
\begin{equation}
	\left[ \min \left(r_\text{min}, \dfrac{r_\text{min}}{1-\gamma}\right), ~
	\max \left(r_\text{max}, \dfrac{r_\text{max}}{1-\gamma}\right)
	\right]
	\label{eq:minmaxQ}
\end{equation}
The \texttt{min}/\texttt{max} operations are required for terminal states. 

Analogously, if the reward only depends on the current state and the agent stays in a non-terminal state $s$ forever, because action $a=\pi(s)$ does not lead to a change in states, then $R^\pi=\frac{r}{1-\gamma}$.
This transfers to larger loops, e.g.\ if transitions $(s_1, a_1, r_1, s_2)$ to $(s_{n}, a_{n}, r_{n}, s_1)$ are known to be induced by a policy $\pi$,
\begin{equation}
	R_1^\pi = \underbrace{r_1 + \gamma r_2 + ... + \gamma^{n-1} r_n}_{r_L} + \gamma^n r_1 + ... =  \sum_{t}^{\infty} (\gamma^n)^t r_L = \dfrac{r_L}{1-\gamma^n}.
	\label{eq:loops}
\end{equation}

The expected future return for executing an arbitrary action $a_t$ and then following the policy is called Q-value:
\begin{equation}
	Q^\pi(s_t, a_t) = \mathbb{E} \left[ r_t + \gamma \cdot R_{t+1}^\pi\right].
	\label{eq:q_definition}
\end{equation}
The agent's goal is to find the optimal policy $\pi^*$ such that the expected future return is maximized from all states.
This can be achieved by finding (a good approximation to) the Q-function and then choosing the action with highest Q-value in each state.

Based on the definition in~\autoref{eq:q_definition}, Q-values can be estimated directly from empirically sampled return values -- so-called Monte Carlo estimates.
This method is known to introduce high variance into the estimates though, because the return can exhibit high variation over long trajectories.

\paragraph*{Temporal Difference Learning}~\\
A popular alternative to Monte Carlo estimates for Q-learning is temporal difference (TD) learning.
Given a transition $(s_t,a_t,r_t,s_{t+1}, \mathfrak{t}_t)$, target Q-values are computed based on the current state value estimate for state $s_{t+1}$:
\begin{equation}
	Q_\text{target}(s_t,a_t) = r_t + \begin{cases}
		0,& \text{if }\mathfrak{t}_t\text{, i.e.\ }s'\text{ is terminal }\\
		\gamma \cdot \mathcal{Q}(s_{t+1}, \pi(s_{t+1})),& \text{else}\\
		\end{cases}.
\label{eq:TD}
\end{equation}

In small settings with finitely many states and actions, \textit{tabular} Q-learning can be applied in which each state-action value Q is represented as one entry in a lookup table.
To update such a Q-function, each Q-value $Q(s,a)$ can be replaced by the target value $Q_\text{target}(s,a)$ directly.

In continuous state or action spaces, \autoref{eq:TD} can be used with function approximation instead.
One of the most popular function approximators for Q-functions are neural networks:
In deep Q networks (DQN), a single network is trained to take states as an input and predict one Q-value for each possible action~\citep{mnih2015human}. 
For continuous actions, an actor-critic architecture called deep deterministic policy gradient (DDPG, ~\citet{lillicrap2015continuous}) can be used: 
The critic is represented by one network that computes the Q-value for a given state-action pair.
The network is trained by minimizing the following loss over data from $N$ transitions:
\begin{equation}
\mathcal{L}_\text{critic} = \frac{1}{N} \sum_{i=0}^N (Q_\text{target}(s_i, a_i) - \mathcal{Q}(s_i, a_i))^2
\label{eq:critic_loss}
\end{equation}
where $\mathcal{Q}$ is the current critic estimate and $Q_\text{target}$ is computed using~\autoref{eq:TD}.

These Q-estimates are used as a training signal for the actor, which is a neural network that represents the policy.

Iteratively updating a function based on its own current estimates is called \textit{bootstrapping}.
Temporal Difference learning is known to introduce less variance than Monte Carlo estimates but higher bias.
Note that bootstrapping is actually only applied in the case of non-terminal states (i.e.\ in the second line of the equation). We will refer to states that do not require bootstrapping to estimate a Q-value as \textit{anchors}.

\paragraph*{Experience Replay}~\\
Both DDPG and DQN use off-policy data, i.e.\ they store past experience in a replay memory and update their networks based on this experience, even if the policy $\pi$ has changed since the data was collected.
Experience is represented by transitions $(s_t, a_t, r_t, s_{t+1}, \mathfrak{t}_t)$, where $s_t$ is the state from which action $a_t$ was taken, $r_t$ is the reward received after reaching state $s_{t+1}$, $\mathfrak{t}_t$ is an indicator for whether or not $s_{t+1}$ is a terminal state.

It is insightful to note that any replay memory only contains a finite number of transitions, that all network updates in DQN and DDPG are derived from, even for continuous state-action spaces.
The original reasoning behind replay memories and experience replay was to break dependencies between transitions~\citep{mnih2015human}, which is important for most function approximation schemes.
We will therefore keep the principle of random selection of transitions for our learning process, but at the same time we will make use of additional information that a graph perspective can provide and would be lost otherwise.

\section{Related Work}
\paragraph*{Instabilities in Reinforcement Learning: the Deadly Triad}~\\
Reinforcement Learning (RL) has been known to be instable even with linear function approximation for more than 20 years~\citep{baird1995residual}.
RL with function approximation, bootstrapping and off-policy learning has been called \textit{deadly triad} by~\citet{sutton2018reinforcement} because it is even more prone to divergence.
Deep RL methods within the deadly triad however seem to exhibit soft divergence rather than unbounded divergence; i.e.\ they often under- or overestimate Q-values but do not reach floating point \texttt{NaN}s~\citep{van2018deep}.
While some researchers work towards provably stable methods (e.g. \citep{ghiassian2018online, degris2012off}),
our work builds on research towards understanding and counteracting soft divergence in deep RL.
In particular, divergence due to an algorithm being in the deadly triad can be counteracted by decreasing the impact of each of the triad properties:

Different networks for \textit{function approximation} and update schemes have been linked to convergence:
\citet{fu2019diagnosing} found large neural networks with compensation for overfitting to be beneficial for learning stability. 
A target network is a second function approximator that is only updated slowly or periodically~\citep{mnih2015human}.
Its values are therefore more stable and lead to more stable target Q-values in temporal difference learning.
Besides, a second network can help to counteract maximization bias in Q-learning~\citep{van2016deep}.
Also other methods that delay~\citep{fujimoto80addressing} or average target values~\citep{anschel2017averaged} have been shown to stabilize learning.
\citet{achiam2019towards} theoretically link generalization properties of the Q-function approximator to the stability of learning. 
We empirically confirm and provide further intuition about this effect in~\autoref{sec:linking}.

In policy gradient methods, reducing the impact of \textit{off-policy} data has been beneficial for stability, e.g.\ by mixing on- and off-policy~\citep{NIPS2017_6974} or by constraining the gradient update through a proximity term~\citep{touati2020stable}.
Also in DQN and DDPG, restricting the action space to achieve lower levels of off-policy data have been explored~\citep{fujimoto2019off}.
Constrained action selection when computing the target Q-values can also stabilize deep RL~\citep{kumar2019stabilizing}.\\
\citet{kumar2020discor} show that the interaction of off-policy learning and bootstrapping can lead to cases where a state is visited frequently and yet its incorrectly estimated Q-value is not updated because the state that the target value depends on is not visited.
They refer to this phenomenon as 'lack of corrective feedback', which we will get back to in our analysis in the next section.
From their observation, they derive a re-weighting of transitions from the replay buffer that is supposed to mitigate this issue.
The full version of our method, using zero actions, will be able to improve performance with such tail ends of data distributions without downweighting the associated transitions and without an additional error model and without constraining the action selection.\\
Off-policy corrections in general are not entirely understood yet:
On the one hand, they may also have adverse effects, e.g. as reported by \citet{hernandez2019understanding} for SARSA.
On the other hand, \citet{fedusrevisiting} found that counter-intuitively, n-step return updates which are not corrected for policy differences are beneficial in off-policy deep RL despite being theoretically ungrounded.

Standard Q-learning uses \textit{bootstrapping} as in \autoref{eq:TD} to estimate a Q-function.
Alternatives to bootstrapping include fixed-horizon temporal difference methods~\citep{de2019fixed}
and finite-horizon Monte Carlo updates, in which a Q value is estimated based on observed Returns from each state.
While the resulting estimator for the Q function has low bias, it comes with high variance.
Combining TD learning with eligibility traces of different lengths, a spectrum of methods between TD and Monte Carlo methods can be spanned~\citep{sutton2018reinforcement, precup2000eligibility}, also in a deep learning setting~\citep{munos2016safe, mnih2016asynchronous, amiranashvili2018analyzing}.\\
Monte Carlo updates can be seen as a special case of graph-perspective: data from full episodes is used to derive updates along a trajectory. 
Similarly to these methods, the lower bounds in our case propagate information along full trajectories.
However, we do not apply return values as high-variance targets but use them to derive a single lower bound each target Q-value (\autoref{eq:TD}) instead.

Our methods uses the full amount of off-policy data that is available, manages to use function approximation without target or double networks and target Q-values are computed based on bootstrapping. However, these target values are constrained by bounds derived from a graph perspective on the training data.
In the following two paragraphs, we will review other works that make use of a graph or trajectory perspective on the training data as well as methods introducing constraints in Q-learning.

\paragraph*{Graph Perspective on Training Data}~\\
While return-based methods such as Monte Carlo estimates for Q-values take an implicit graph perspective, there is related work building explicit graphs:\\
Episodic backward updates are classical TD updates that are executed along trajectories in reverse order, such that information is quickly propagated through consecutive states~\citep{lee2019sample}.
To prevent errors from consecutive updates of correlated states, a diffusion coefficient is introduced.\\
\citet{zhu2019episodic} take a full graph perspective on the agent's experience: 
using a learned state embedding, episodes with shared states are identified and can benefit from inter-episode information, i.e. the algorithm can combine multiple trajectories from experience.
State embeddings have also been combined with k-nearest neighbors as a method to estimate Q-values for unseen states~\citep{blundell2016model}.
\citet{corneil2018efficient} use a network model to map states to an abstract tabular model where planning can be easily applied.
In our approach, we also use a graph perspective but without a learned embedding inter-episodic information is only exchanged if the exact same state is revisited (up to floating point precision).

\paragraph*{Constrained Q-learning}~\\
Q-learning can be stabilized by introducing constraints on the change in either target values or network parameters~\citep{durugkar2018td,ohnishi2019constrained}.
However, constraining change rates in a learning system may also limit the rate at which an agent can improve.

\citet{he2019learning} suggest to apply both upper and lower bounds to target Q-values, which are based on the current Q-estimate and therefore additional multiple forward passes in each update step. Because these bounds are based on the current Q-estimate, they need not be correct in general.
In contrast, we will derive correct lower bounds for $\pi^*$ in near-deterministic settings and show that incorrect empirical bounds can have adverse effects on the learning process.

\citet{tang2020self} offers the intuition that lower bounds encourage the algorithm to focus on the best actions so far and thereby speed up learning.
This idea is in line with \citet{zhang2019asynchronous} who introduce a separate replay buffer that only holds the best episodes and empirically improves learning performance on a range of simulated continuous control tasks.

\section{Linking Data Graph Structure to Soft Divergence}
\label{sec:linking}
Despite the continuous state-action space, the networks in DDPG are updated based on a finite set of transitions from the replay memory.
It is therefore possible to take a graph perspective on this data:
A transition $(s,a,r,s',\mathfrak{t})$ can be seen as an edge between the nodes corresponding to states $s$ and $s'$ (which is terminal iff indicated by $\mathfrak{t}$); and can be annotated with action $a$ and reward $r$.
Any hashing function can be used to encode nodes and detect if the same node is revisited. 
This is not supposed to introduce any discretization beyond the limits of precision.
We refer to the resulting directed graph as \textit{data graph} (see \autoref{fig:teaser} for an illustration).

\bigskip

The structure of a data graph can be linked to soft divergence in deep Q-learning as the following example demonstrates:\\
We examine a task where an agent can maneuver in a 2D continuous state space with 2D actions such that adding state and action yields the next state $s_{t+1} = s_t + a_t$.
For each step, the agent receives a reward of $-1$ and $0$ at the terminal state.\\
Let's assume a DDPG-like critic network is trained to find an approximation to the Q-function for this problem.
We chose two layers with 4 hidden states, ReLU activations (except on the output) and Xavier-initialization.

\begin{figure}
	\centering
    \begin{minipage}{7cm}
        full data graph:\\
        \hspace{0.3cm} \begin{tikzpicture}[scale=0.7]
\def\xdist{0.7}
\def\curvature{20}
\node[circle, draw, double](0) at (0,0) {{\small 0}};
\node[circle, draw](1) at (-\xdist,1) {{\small 1}};
\node[circle, draw](2) at (\xdist,1) {{\small 2}};
\path
(1) edge[->, bend left=\curvature] (2) 
(1) edge[->] (0) 
(2) edge[->, loop, looseness=3.5] (2) 
(2) edge[->, bend left=\curvature] (1);
\node[draw,minimum height=1.8cm,minimum width=2cm] at (0,0.65) {};
\end{tikzpicture}
    
    	\vspace{0.3cm}
    
        exemplary transition subsets:\\
        \hspace{0.3cm} \fbox{\begin{tikzpicture}[scale=0.75]
\def\xdist{0.7}
\def\curvature{20}
\node[circle, draw, double](0) at (0,0) {{\normalsize0}};
\node[circle, draw](1) at (-\xdist,1) {{\normalsize1}};
\node[circle, draw](2) at (\xdist,1) {{\normalsize2}};
\path
(1) edge[->, bend left=\curvature, draw=sns2, very thick] (2) 
(1) edge[->, draw=sns1, very thick] (0) 
(2) edge[->, bend left=\curvature, draw=sns2, very thick] (1);
\end{tikzpicture}}
        \hspace{0.3cm} \fbox{\begin{tikzpicture}[scale=0.75]
\def\xdist{0.7}
\def\curvature{20}
\node[circle, draw, double](0) at (0,0) {0};
\node[circle, draw](1) at (-\xdist,1) {1};
\node[circle, draw](2) at (\xdist,1) {2};
\path
(1) edge[->, bend left=\curvature, draw=sns3, very thick] (2) 
(1) edge[->, draw=sns1, very thick] (0) ;
\end{tikzpicture}}
        \hspace{0.3cm} \fbox{\begin{tikzpicture}[scale=0.75]
\def\xdist{0.7}
\def\curvature{20}
\node[circle, draw, double](0) at (0,0) {0};
\node[circle, draw](1) at (-\xdist,1) {1};
\node[circle, draw](2) at (\xdist,1) {2};
\path
(1) edge[->, bend left=\curvature, draw=sns4, very thick] (2) 
(2) edge[->, bend left=\curvature, draw=sns4, very thick] (1);
\end{tikzpicture}}
    \end{minipage}
	\hspace{0.3cm} 
    \begin{minipage}{6.5cm}
        \hspace{0.5cm}
        \includegraphics[width=\linewidth]{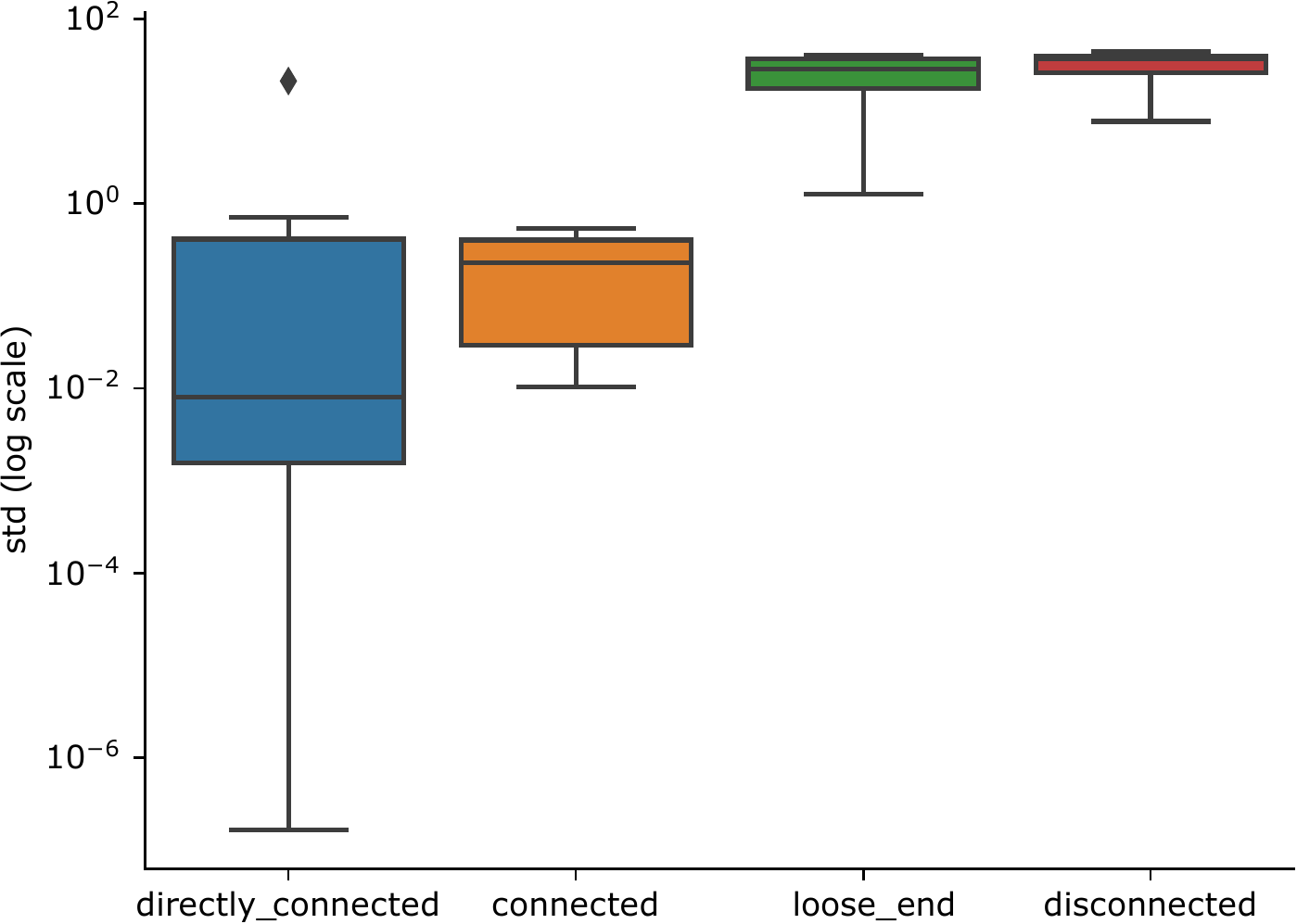}
    \end{minipage}
    \caption{Educational example with four transitions and three states (state 0 is terminal).
        We characterize transitions based on the graph structure:
        (in-)directly connected to a terminal state (blue, orange); loose ends (green) and disconnected but infinite paths (red).
        The right plot illustrates the standard deviation over predicted Q-values for each type of transition from all 15 possible subsets of the educational example.
    }
    \label{fig:education}
\end{figure}
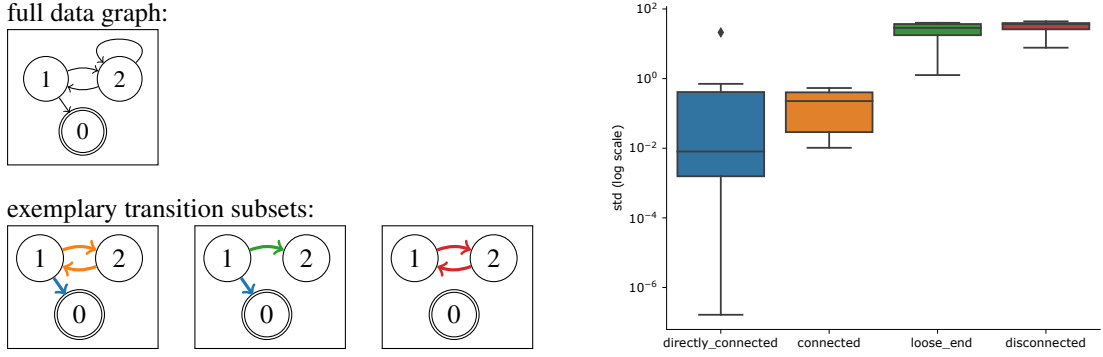

All network updates are solely derived from the replay memory, which is filled with any subset of the four transitions shown in~\autoref{fig:education} and then fixed for offline policy evaluation on the known state-action pairs. 
There are $2^4=16$ different subsets of experience with different data graph structures; one of which is empty and therefore ignored.
For the remaining 15 cases, we have trained the critic network with ten thousand training epochs consisting of all available transitions.
The states were assigned 2D coordinates as follows:  $s_0 = [0,0]$, $s_1 = [-1,1]$, $s_2 = [1,1]$.\\
The training procedure was repeated with 10 random seeds that were drawn uniformly from $[0,1000]$.
No actor network was trained and instead, the known action from the replay memory with highest associated Q-value was chosen to compute the Q-targets.

Confirming the finding in \citet{van2018deep}, no \textit{unbounded} divergence occurred (which would cause floating point \texttt{NaN}s).
However, we found occurrences of \textit{soft} divergence, i.e.\ Q-values beyond the realizable range as given by \autoref{eq:minmaxQ}.\\
For further analysis we compute the standard deviation of Q-values that were estimated over different random seeds as a measure of soft divergence: 
if Q-learning for a transition converges, all Q-values should be identical and thus have a standard deviation close to zero. 
The more soft divergence occurs however, the larger the standard deviation becomes. 
Even if all trials diverge, it is highly unlikely that the resulting Q-values are identical.

\bigskip

Evaluating the distribution of standard variations reveals a link between the structure of the Q-graph and soft divergence:

\begin{enumerate}
	\item 
	Transitions $(s,a,r,s',\mathfrak{t})$ where $s'$ is terminal are referred to as \textit{directly connected}.
	Their Q-values are estimated almost perfectly, because Q-learning is reduced to supervised learning in these cases (cf.\ ~\autoref{eq:TD}).
	
	\item 
	Transitions that end in a non-terminal state from which a terminal state is reachable are referred to as \textit{connected}.
	Their Q-estimates exhibit only slightly more variance than the directly connected transitions. Presumably the reachable terminal state still acts as an anchor for the Q-value (as long as all transitions on the path are regularly used for updates).
	In line with this hypothesis, the two following categories that do not have an anchor show significantly more variance in their predictions:
	
	\item 
	If no terminal state is reachable from $s'$ and there is no infinite path from $s'$, the transition is referred to as a \textit{loose end}.
	These transitions occur for instance at the end of each episode in episodic learning setups, when the agent does not succeed but is reset to a starting position.
	It is insightful to note that Q-values for such transitions are conceptually ill-defined in tabular Q-learning where a state without successors would be defined as terminal.
	For non-terminal states, a Q-value could be determined under the assumption that further transitions exist (and just have not been experienced yet), but then the Q-value is estimated using bootstrapping from another Q-value that has never been explicitly updated.
	This phenomenon is one example for what has been referred to as a lack of corrective feedback~\citep{kumar2020discor}.
	In other words, the estimate depends only on network initialization and generalization from data for other state-action pairs; cf. also \citet{achiam2019towards} who analyze the theoretical link between approximator generalization properties and learning stability.

	\item 
	Transitions are referred to as \textit{disconnected} if no terminal state is reachable from $s'$ but there exists at least one infinite path from $s'$. 
	In applications of reinforcement learning, these transitions occur frequently, e.g. when the agent gets stuck in a non-terminal state.\\
	Disconnected transitions caused the highest variance in Q-estimates.
	In contrast to loose ends however, the Q-value for these transitions is well-defined under the assumption that all possible transitions are known and can even be computed analytically (cf.\ ~\autoref{eq:loops}).
\end{enumerate}

\bigskip

We draw the following conclusions from this introductory experiment:
\begin{enumerate}
	\item There is a clear link between the data graph structure and soft divergence; even for a static replay memory, a simple restricted policy and very few transitions.
	
	\item Episodic tasks which create loose ends lead to an ill-posed estimation task which can only rely on generalization capabilities of the Q-function approximator.
	
	\item Disconnected transitions pose a well-defined estimation problem and yet they cause the highest variance in Q-estimates in our experiment.
\end{enumerate}

Our method, which will be presented in detail in the following section, extracts the largest possible subgraph for which exact Q-values can be computed under the assumption that all possible transitions are known.
These Q-values represent a lower bound for the Q-value in the original continuous learning problem and enforcing them in temporal difference learning can stabilize learning.
We will show empirically that, besides further effects, this reduces the variance of predicted Q-values also for a more realistic peg-in-hole continuous control task.

\section{Q-graph bounded Q-learning}
\label{sec:bounds}
Building on the insights from \autoref{sec:linking}, we select the largest set of transitions from the \textit{data graph} for which exact Q-values can be computed under the assumption that the resulting graph is complete (i.e.\ that all possible transitions and all states are included).
That is, we extract all transitions from the data graph except for loose ends.
Formally, this induces a smaller finite MDP for which the associated Q-function can be computed using tabular Q-iteration with guaranteed convergence due to its contraction property.
Our method is agnostic to the algorithm that computes these Q-values, so for instance it is also possible to solve the linear equation system for a sparse transition matrix.
In any case, the computational overhead to compute these Q-values depends on the number of transitions in the replay memory, but it is independent of the input dimensionality.
We annotate the subgraph of the data graph with the resulting Q-values and refer to it as \qgraph.
One possible implementation of a \qgraph{} is illustrated in \autoref{alg:graph} in \autoref{app:pseudocode}.

In many settings, there are known \textit{zero actions} $a_z$ that do not change the agent's state at all, e.g.\ moving by 0 units or applying 0 force.
If those are applicable in all states, it may be possible to add a self-loop to every single node in the data graph. 
This effectively eliminates all loose ends and turns them into disconnected states, in other words it allows the \qgraph{} to contain all transitions from the data graph and compute their exact Q-values for the simplified MDP.

\paragraph*{\qgraph{} Values as Lower Bounds}~\\
In general, the original MDP contains more states or transitions than the \qgraph{}.
Then, the Q-values do not transfer to the original MDP as a correct solution but can be used as lower bounds for Q-values in the original MDP.

Assume w.l.o.g. that at least two transitions $(s_0,a_1,r_1,s_1)$ and $(s_1,a_2,r_2,s_2)$ are known and part of the \qgraph{} $\mathcal{G}$ with associated Q-values $\mathcal{Q}_\mathcal{G}$ for the associated simplified discrete MDP.
Since Q-values for all transitions in $\mathcal{G}$ can be computed exactly using Q-iteration, the Bellman optimality equation applies:
\begin{equation}
\mathcal{Q}_{\mathcal{G}}(s_0, a_1) = r_1 + \max_{a \in \mathcal{G}_{s_1}} \mathcal{Q}_\mathcal{G}(s_1, a) 
\label{eq:Q_G}
\end{equation}
where $\mathcal{G}_{s_1}$ denotes all actions on out-going edges from $s_1$.

In the original MDP with potentially continuous state and action spaces, unseen states and transitions may exist.
Still, in deterministic MDPs, the Q-value for the full MDP is lower bounded due to the $\max$ operation and the fact that the available actions in the \qgraph{} ($\mathcal{G}_{s_1}$) are a subset of those in the continuous action space $\mathcal{A}$:
\begin{align}
	\mathcal{Q}(s_0, a_1) =&~ r_1 + \max_{a \in \mathcal{A}} \mathcal{Q}(s_1, a) \label{eq:qgraph_max_F} \\
	                     \geq&~ r_1 + \max_{a \in \mathcal{G}_{s_1}} \mathcal{Q}(s_1, a)  \label{eq:qgraph_max_G}\\
	                     =&~	\mathcal{Q}_{\mathcal{G}}(s_0, a_1)
	                     \label{eq:qgraph_max}
\end{align}

Thus, each Q-value for a transition in our \qgraph{} $\mathcal{G}$ represents a lower bound of the Q-value for the same transition in the original MDP on continuous state and action spaces.
In contrast to the prior work in~\citet{he2019learning}, these lower bounds do not depend on the current Q-estimate but hold for the optimal Q-value in general.

Note that the $\max$ operation in~\autoref{eq:qgraph_max_G} operates on a discrete space and can thus be computed by a simple look-up and comparison of all known transitions from $s_1$.
To evaluate~\autoref{eq:qgraph_max_F} in a continuous space, e.g. for temporal difference learning as in~\autoref{eq:TD}, the maximization is re-written using the currently estimate of the optimal policy $\pi^*_Q(s_1)$:
\begin{equation}
	\max_{a \in \mathcal{A}} \mathcal{Q}(s_1, a) = \mathcal{Q}(s_1, \pi^*_\mathcal{Q}(s_1))
\end{equation}
In the DDPG setting and all our empirical evaluations, $\pi^*_\mathcal{Q}$ is represented by the actor network that is trained to maximize $\mathcal{Q}$.

\bigskip

For non-deterministic dynamics, potentially less tight bounds can be established under additional assumptions:
If for any state and any series of actions $\mathfrak{A}$, the empirical return $R$ that an agent can observe when following $\mathfrak{A}$ from $s$ differs by at most $\delta$, then all Q-values from the simplified MDP apply as lower bounds with margin $\delta$:
\begin{equation}
	\mathcal{Q}(s,a) \geq \mathcal{Q}_\mathcal{G}(s,a) - \delta
	\label{eq:nondet}
\end{equation}

Since non-deterministic environments are quite common and $\delta$ may not be known, we will additionally evaluate the empirical performance of our method under violation of the determinism assumption.

\paragraph*{\qgraph-bounded Q-learning}~\\
\label{par:qgraph-bounded-qlearning}
Bounds on Q-values, for instance those computed in~\autoref{eq:qgraph_max}, can be enforced in temporal difference learning by modifying target Q-values~\autoref{eq:TD} as follows:
\begin{equation}
	\mathcal{Q}_\text{target}(s_t,a_t) = \max \left( \text{LB}_t, r_t + \begin{cases}
	0,& \text{if }s'\text{ is terminal }\\
	\gamma \cdot \mathcal{Q}(s_{t+1}, \pi(s_{t+1})),& \text{else}\\
	\end{cases} \right)
	\label{eq:bounded_td}
\end{equation}
where LB$_t$ is a lower bound; e.g.\ the Q-value for the same transition from the \qgraph{} $\mathcal{Q}_\mathcal{G}(s_t,a_t)$.
If another lower bound is known, e.g.\ based on a bounded reward as in~\autoref{eq:minmaxQ}, LB can be the maximum over all available bounds.
Analogously, upper bounds UB could be applied using the $\min$ operation.

We refer to this method of enforcing Q-values from $\mathcal{G}$ in the target values for temporal difference learning as \qgraph-\textit{bounded Q-learning}.
When the Q-function  $\mathcal{Q}$ is represented by a function approximator, e.g.\ a neural network in DDPG, it is defined for a continuous state and action space.
While training however, the Q-targets are constrained by bounds derived from the \qgraph{}-based $\mathcal{Q}_\mathcal{G}$-values on a discrete domain.

If a state-action pair is not associated with a lower bound, i.e.\ loose ends or transitions leading to such, can be used as usual in \autoref{eq:TD}, i.e.\ without clipping of their target value.
If coincidentally no bounds are violated, our method reduces to vanilla DDPG.
A full training step is illustrated as pseudocode in~\autoref{alg:train} in~\autoref{app:pseudocode}.

\section{Experimental Results}
We evaluated the core of our method on a classical toy example for convergence issues in value learning in~\autoref{subsec:exp-baird}.

Additionally we ran a series of experiments on a continuous control problem (\autoref{subsec:exp-setup}) to evaluate performance in terms of sample efficiency and robustness to hyperparameters in~\autoref{subsec:exp-sampleEff-robustness}.
In~\autoref{subsec:exp-var-preds}, we verify that the outcome on the continuous control problem is in line with the insights about soft divergence from our introductory example in~\autoref{sec:linking}.
We further examine the impact of zero actions and different types of upper and lower bounds on Q-values (\autoref{subsec:exp-baselines}) as well as the method's interaction with limited replay memory capacity (\autoref{subsec:exp-capacity}).
Finally, we empirically asses the impact of non-deterministic transition dynamics in~\autoref{subsec:nondeterminism}.

The usefulness of our method has further been demonstrated on an industrial insertion task in~\cite{hoppe_2020iros}.

\subsection{Baird's Star Problem}
\label{subsec:exp-baird}
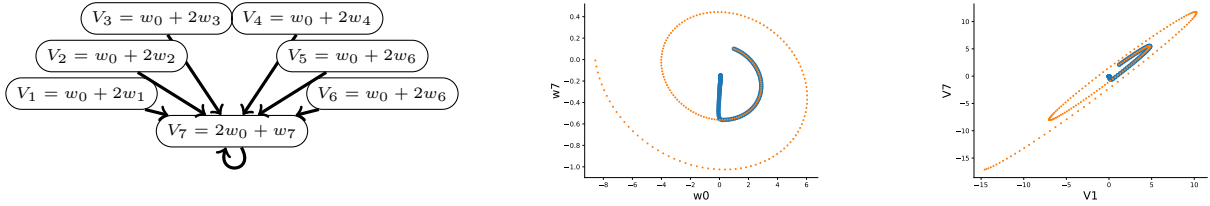
\begin{figure}
	\centering
	
	\raisebox{-0.5\height}{
		\begin{tikzpicture}
			\usetikzlibrary{shapes}
			\def\curvature{20}
			\def\xdist{2}
			\def\ydist{0.5}
			\node[rounded rectangle, draw](V1) at (-\xdist, \ydist) {{\tiny$V_1=w_0+2w_1$}};
			\node[rounded rectangle, draw](V3) at (-0.5*\xdist, 3*\ydist) {{\tiny$V_3=w_0+2w_3$}};
			\node[rounded rectangle, draw](V4) at (0.5*\xdist, 3*\ydist) {{\tiny$V_4=w_0+2w_4$}};
			\node[rounded rectangle, draw](V6) at (\xdist, \ydist) {{\tiny$V_6=w_0+2w_6$}};
			\node[rounded rectangle, draw](V7) at (0,0) {{\tiny $V_7=2w_0+w_7$}};
			\path (V7) edge[->, very thick, loop, looseness=5, in=250, out=300] (V7) 
			(V1) edge [->, very thick] (V7)
			(V3) edge [->, very thick] (V7)
			(V4) edge [->, very thick] (V7)
			(V6) edge [->, very thick] (V7);
			\node[rounded rectangle, draw, fill=white](V2) at (-0.8*\xdist, 2*\ydist) {{\tiny$V_2=w_0+2w_2$}};
			\node[rounded rectangle, draw, fill=white](V5) at (0.8*\xdist, 2*\ydist) {{\tiny$V_5=w_0+2w_6$}};
			\path 
				(V2) edge [->, very thick] (V7)
				(V5) edge [->, very thick] (V7);
		\end{tikzpicture}}
    \hspace{1cm}
    \raisebox{-0.5\height}{\includegraphics[width=4.cm]{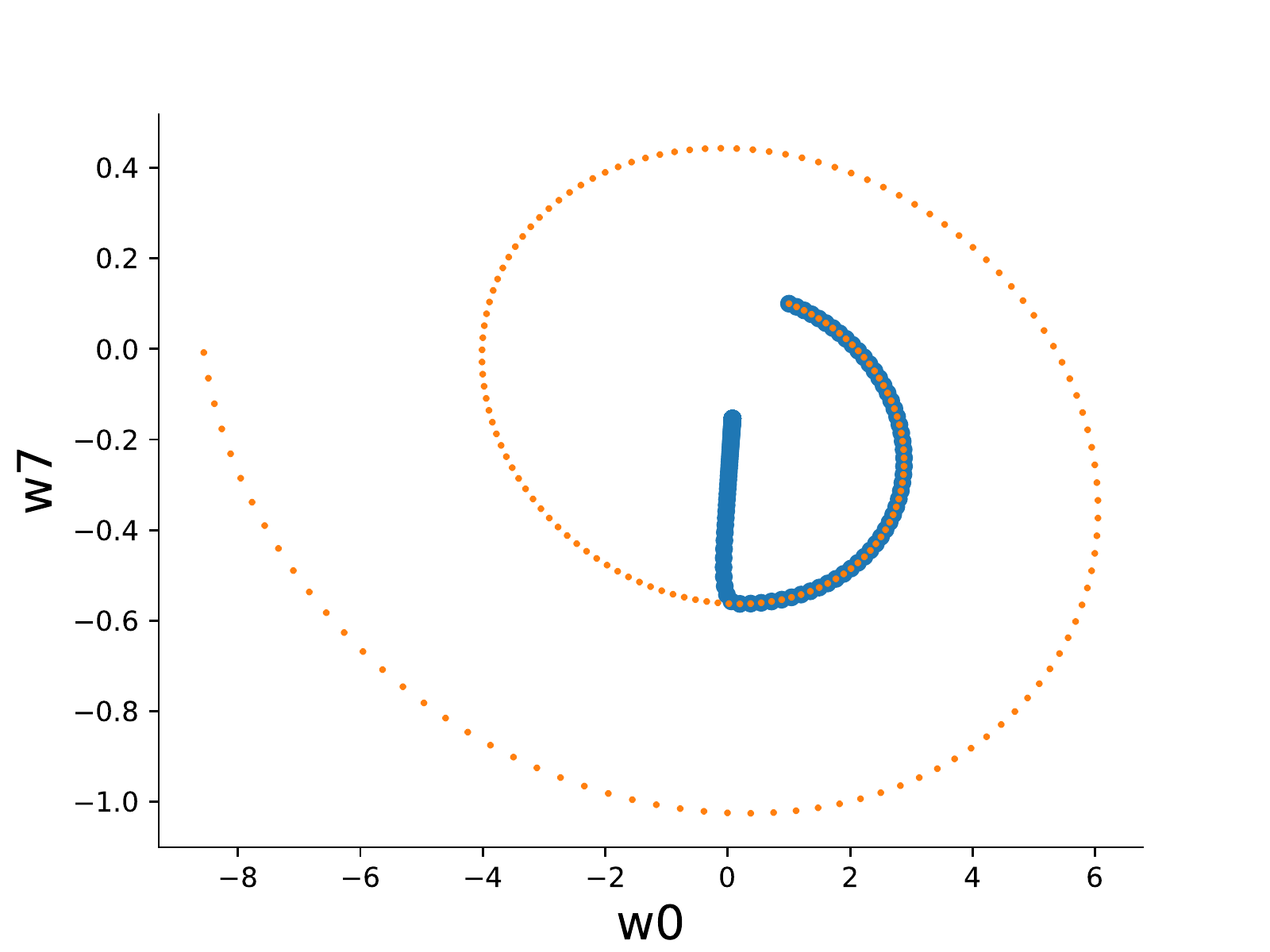}}
    \hspace{1cm}
	\raisebox{-0.5\height}{\includegraphics[width=4.cm]{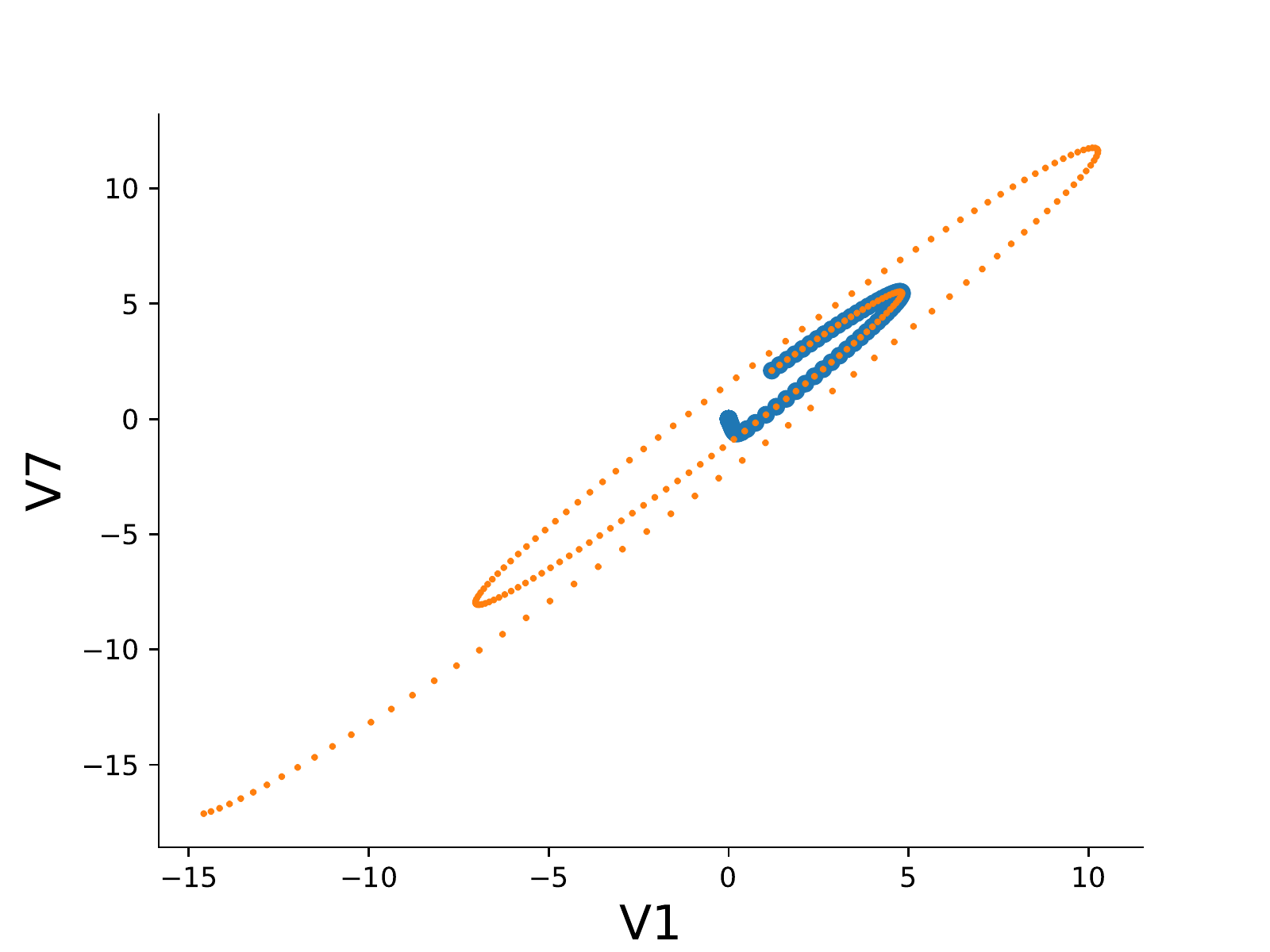}}
	\caption{
        Graph-based bounds lead to the correct solution (blue, solid) on the 7-state star problem after~\citet{baird1999thesis}, for which states and weights spiral out to infinity under vanilla TD learning (orange, dotted).
    }
	\label{fig:star}
\end{figure}

The 7-state star problem (\autoref{fig:star}) was proposed by~\citet{baird1999thesis} to demonstrate convergence issues in value iteration with (linear) function approximation.
The agent receives a reward of zero for each action and thus the correct solution to the problem is to set all weights to zero and obtain state-values of zero.
If all weights are initially positive and $w_0$ larger than the others, this causes oscillatory behavior of both state values and weights.
We reproduced the exact setting and result plots for Figure 4.2 in~\citet{baird1999thesis}.
Applying our graph view to the problem, we can derive a lower bound of zero for $V_7$ because it has a self-loop with reward 0; and thus this lower bound recursively leads to a lower bound of $0 + \gamma V_7 = 0$ for all other states.
These graph-based bounds can be applied in TD learning in analogy to~\autoref{eq:bounded_td} as $V'(s) = \max(LB, r + \gamma V(s'))$.
As a result, our method converges to the correct state values rather than diverging to infinity as~\autoref{fig:star} illustrates.


\subsection{Experimental Setup}
\label{subsec:exp-setup}

\begin{wrapfigure}{r}{3.5cm}
	\begin{center}
		\includegraphics[width=3cm]{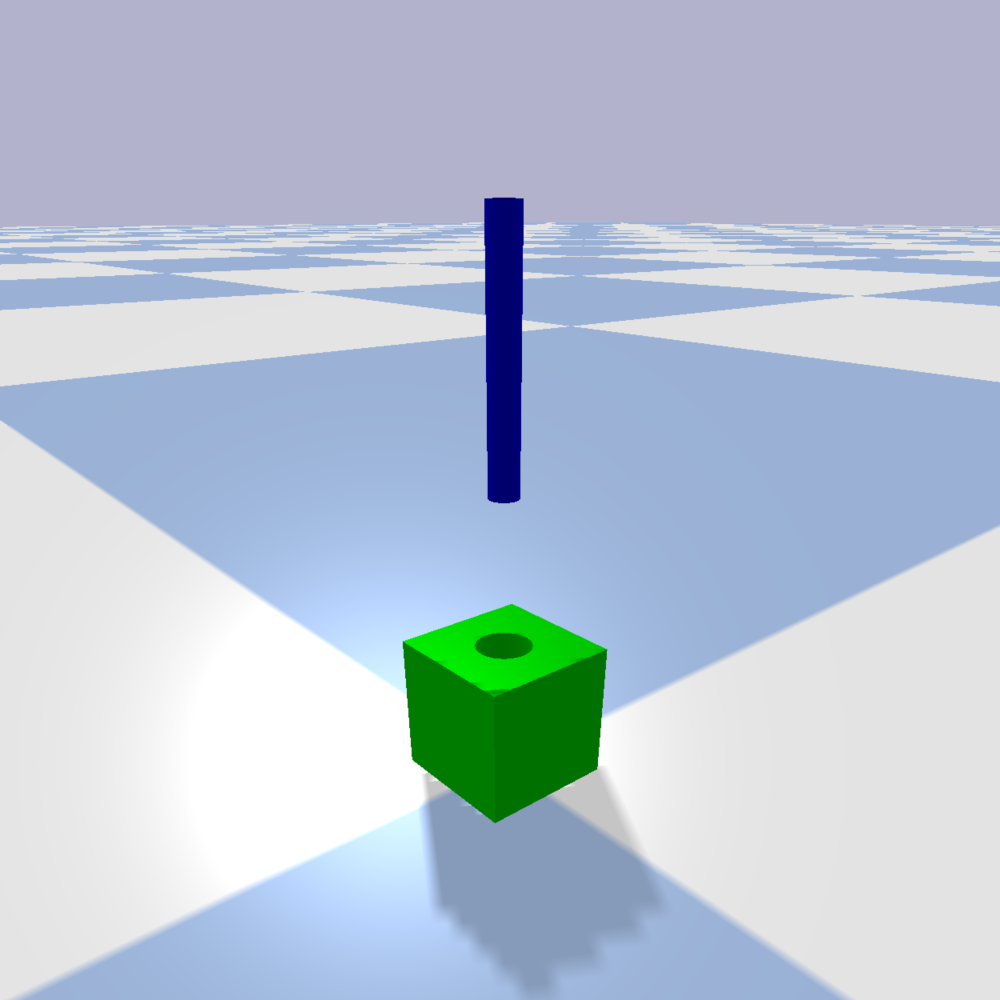}
	\end{center}
	\caption{Simulated Peg-In-Hole task.}
	\label{fig:exp_setup}
\end{wrapfigure}

All further experiments were conducted on a simulated continuous control task.
The environment was implemented using pybullet\footnote{\url{https://github.com/bulletphysics/bullet3}}.
A peg is supposed to be inserted into a green square object, see~\autoref{fig:exp_setup}.
The peg is always upright and velocity-controlled: an action represents the three-dimensional offset to the next position.
The simulation is stepped forward until a stable new position is reached.
The actions are box-constrained to $[-1,1]$ in each dimension which corresponds to a movement of 1cm. 
The green object has a width of 5cm and is within a cubic state space of width 20cm.
The peg has a diameter of 1cm, the hole's diameter is 2cm.
The agent receives a distance-based reward $r=\exp(-\frac{\Delta}{0.03}) - 1$, where $\Delta$ is the Euclidean distance to the goal position in meters.

We use the following instance of a standard DDPG architecture for learning:
The critic network consists of three fully connected layers with 200 nodes each. 
For the inner layers, ReLU activations were used. 
The network was initialized with weights sampled from $\mathcal{N}(\mu=0, \sigma=0.001)$.
The actor network also consists of three fully connected layers with 200 nodes each, but used tanh activations and was initialized from a He-uniform distribution.
All neural networks were implemented using Tensorflow\footnote{\url{www.tensorflow.org}} and optimized using the AdamOptimizer, with 50 training epochs after each episode (i.e.\ 200 agent steps) and up to 15 random mini batches of data per epoch.
No target network was used, since those are known to prolong training and thereby postpone convergence issues but not solve them~\citep{van2018deep}.

\begin{figure}[t!]
    \centering
    \includegraphics[width=0.49\linewidth]{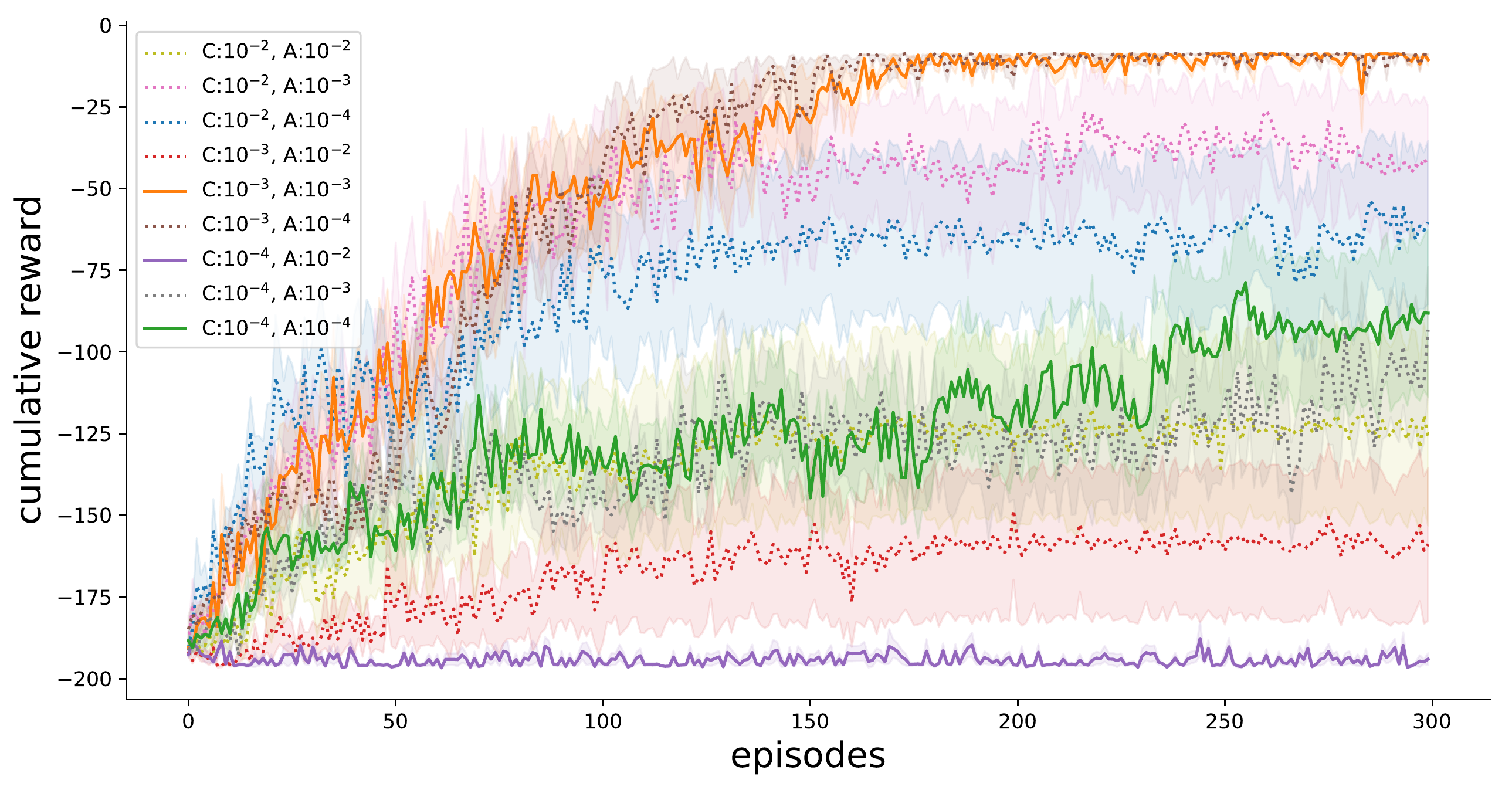}
    \includegraphics[width=0.49\linewidth]{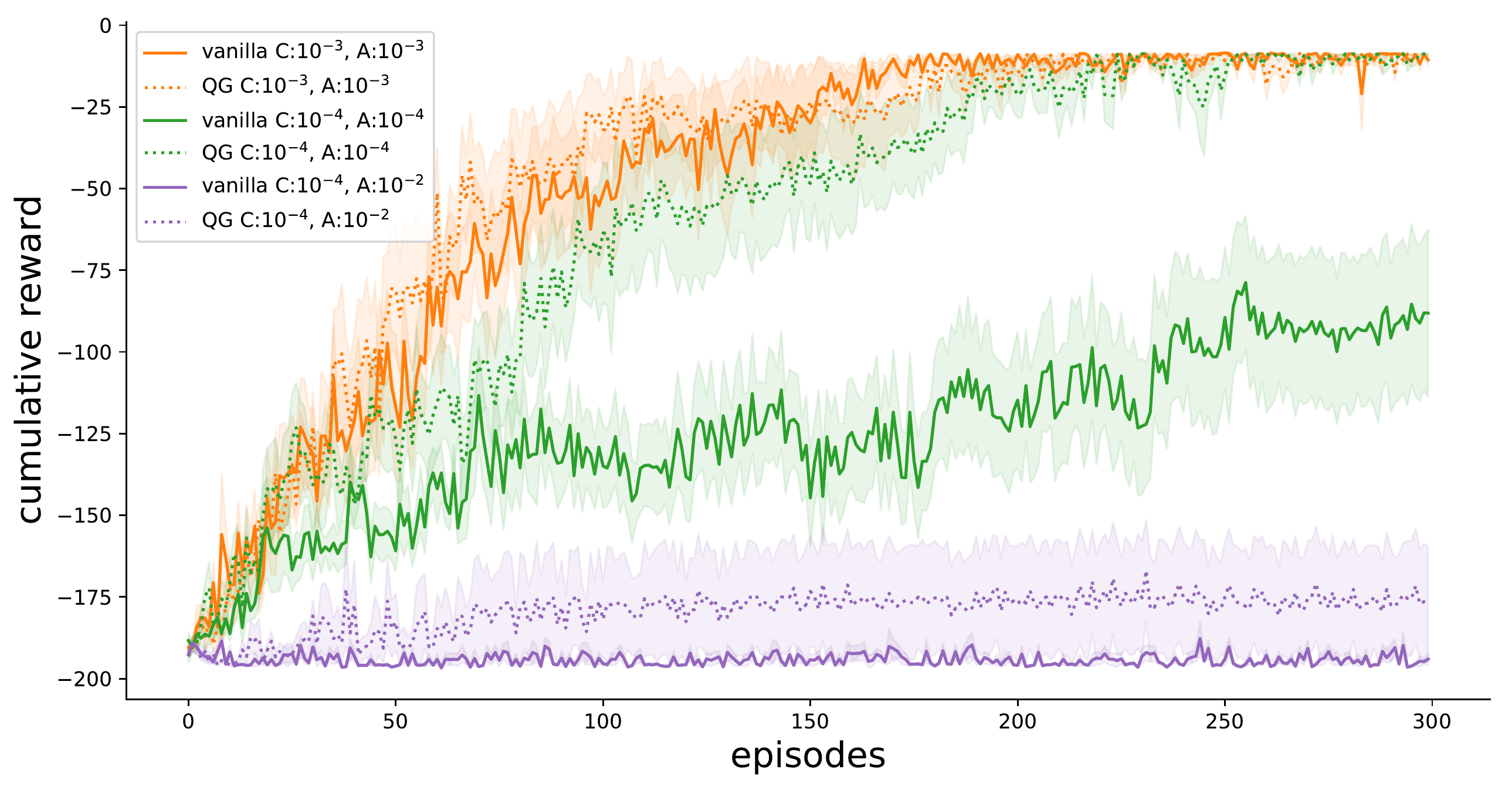}
    \caption{Performance of vanilla DDPG on the full grid of learning rates (left). Three representative parameters were identified (solid lines) and compared to our method ('QG', dotted lines) on the right plot.}
    \label{fig:overview_LR_and_sampleEfficiency}
\end{figure}

We tested vanilla DDPG for 300 episodes on a grid of learning rates for actor and critic in $\{10^{-2},10^{-3},10^{-4}\}$ and chose three sets of hyperparameters for the following experiments that are representative for the spectrum of DDPG performance, see~\autoref{fig:overview_LR_and_sampleEfficiency}.

In all plots with learning curves, the line represents the mean performance over ten runs with different random seeds and the shaded area highlights the standard deviation of the mean estimator, i.e.\ $\frac{\sigma}{\sqrt{n}}$.

\subsection{Sample Efficiency and Robustness to Hyperparameters}
\label{subsec:exp-sampleEff-robustness}
We hypothesized that \qgraph-based lower bounds would correctly limit the range of Q-values which prevents some cases of soft divergence and thereby increases sample efficiency.
We further hypothesized that explicit bounds would barely have any impact in cases when vanilla Q-learning works well, because our method as described in~\autoref{eq:bounded_td} reduces to standard TD learning when no bound is violated.
In other words this implies that \ourmethod{} should never decrease performance.\\
For a first overview, we compared learning curves of \ourmethod{} ('QG') to those of vanilla DDPG in~\autoref{fig:overview_LR_and_sampleEfficiency}.
As expected, \qgraph{}s speed up learning for all examined learning rates.
The effect size varies and is larger for those learning rates that lead to relatively poor performance in vanilla DDPG.
This decreases the gap in performance between different learning rates and can therefore be interpreted as an indicator for increased robustness to hyperparameters.

\subsection{Variance of Predictions}
\label{subsec:exp-var-preds}

\begin{wrapfigure}{r}{6cm}
	\vspace{-1cm}
    \centering
    \includegraphics[width=\linewidth]{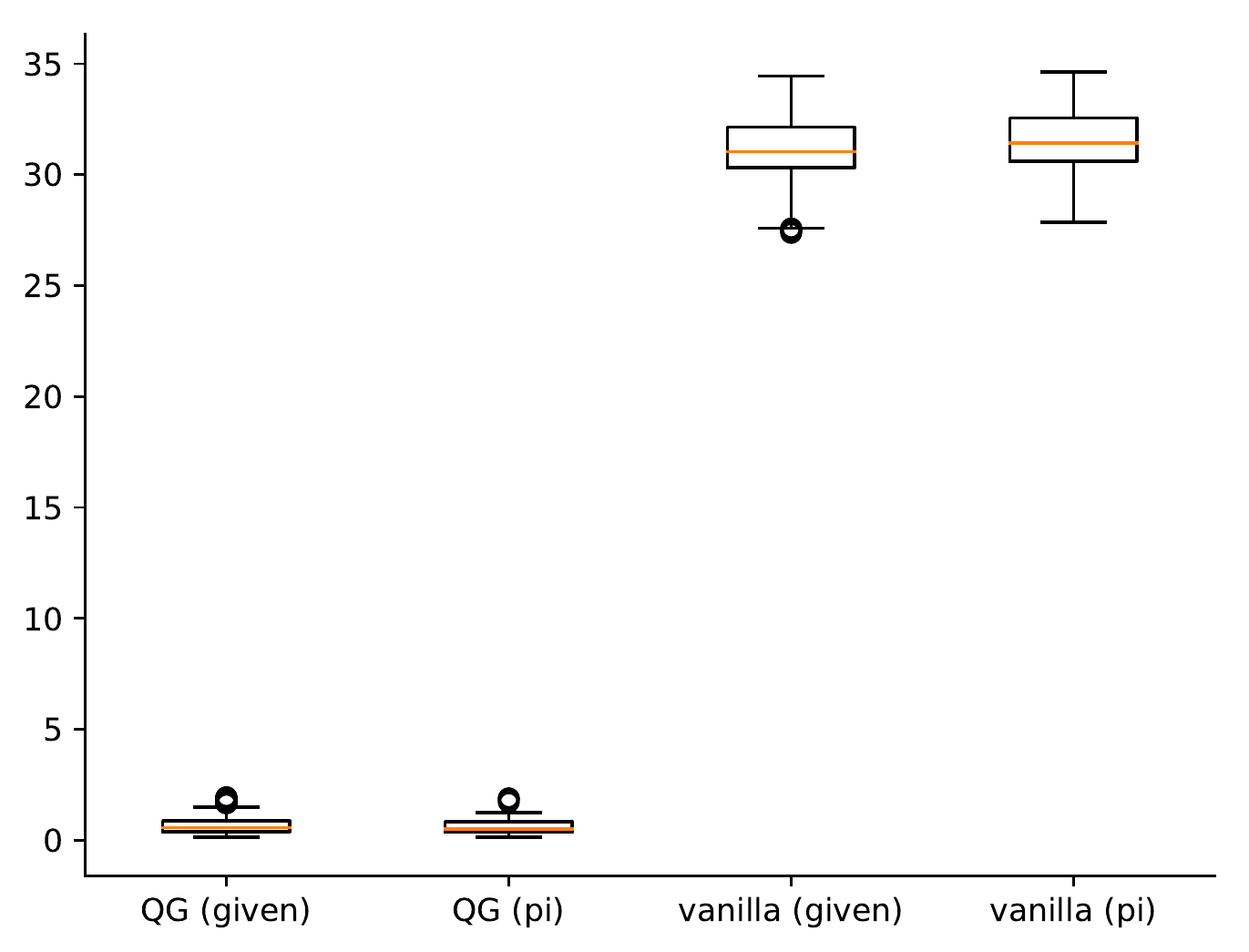}
    \caption{Standard deviation of predicted Q-values.}
    \label{fig:predvar}
    \vspace{-0.3cm}
\end{wrapfigure}

To assess if this increase in performance is due to similar effects as in our educational examples, we evaluated the variance in predicted Q-values at the end of each experiment under the learning rate with largest effect size ($10^{-4}$).
We covered the state space with a regular grid of 27 states and evaluated the learned Q-value for each of these states with a set of eleven given actions ('given') as well as with the action that the actor network suggests for each state ('pi'). \\
For the boxplot in~\autoref{fig:predvar}, we collected the standard deviations over the predicted Q-values for each state-action pair from 10 runs with different random seeds.
The orange line indicates the median value, the box extends from the lower to the upper quartile value, the whiskers cover 1.5 times the inter quartile range and outliers are shown as circles.
The results shows very clearly that \qgraph-runs resulted in significantly less variance for predicted Q-values, indicating that \ourmethod{} does indeed prevent cases of soft divergence. 

\subsection{Further Baselines}
\label{subsec:exp-baselines}
\begin{figure}[t]
    \begin{center}
		\includegraphics[width=0.49\linewidth]{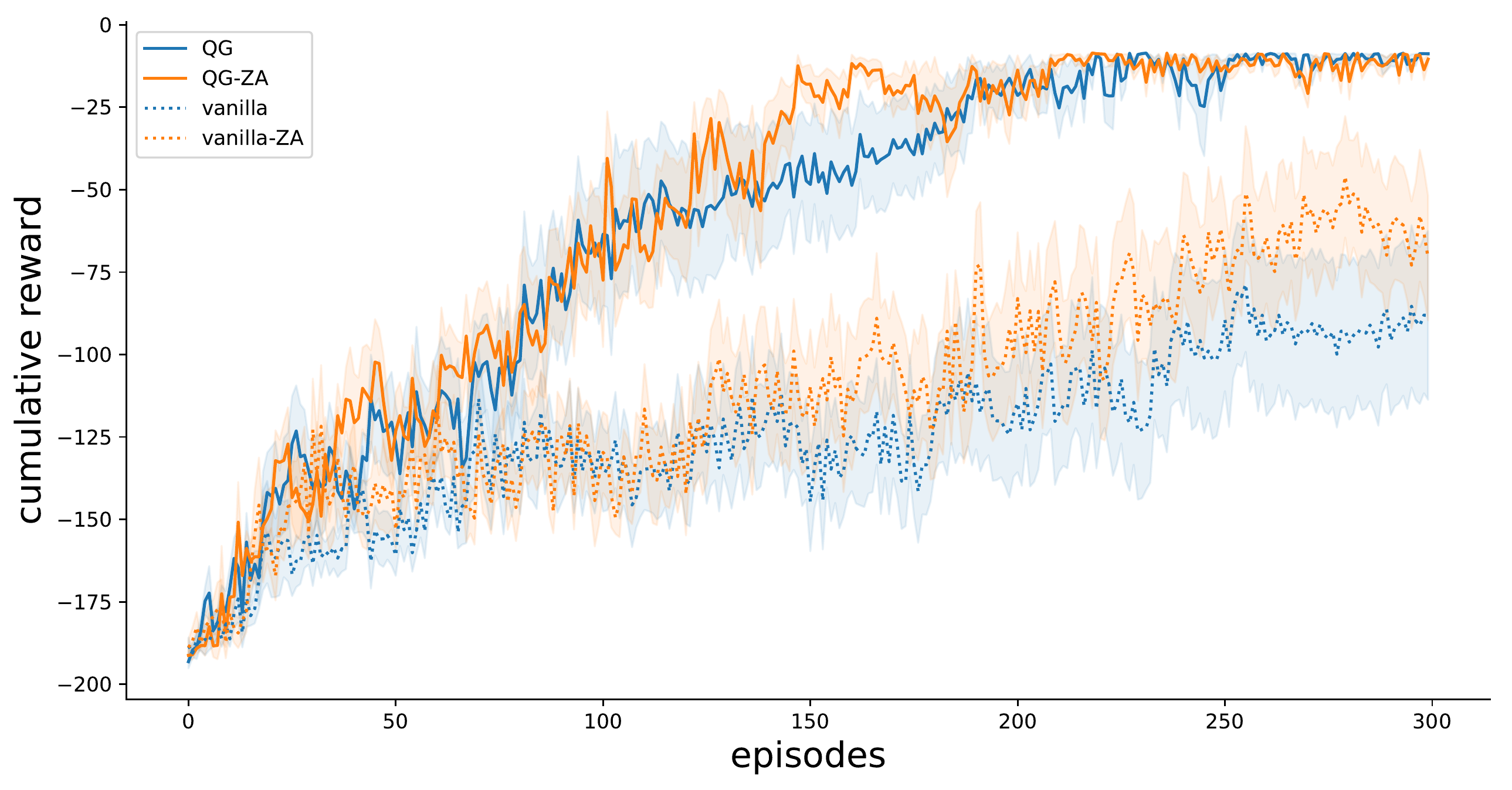}
		\hfill
		\includegraphics[width=0.49\linewidth]{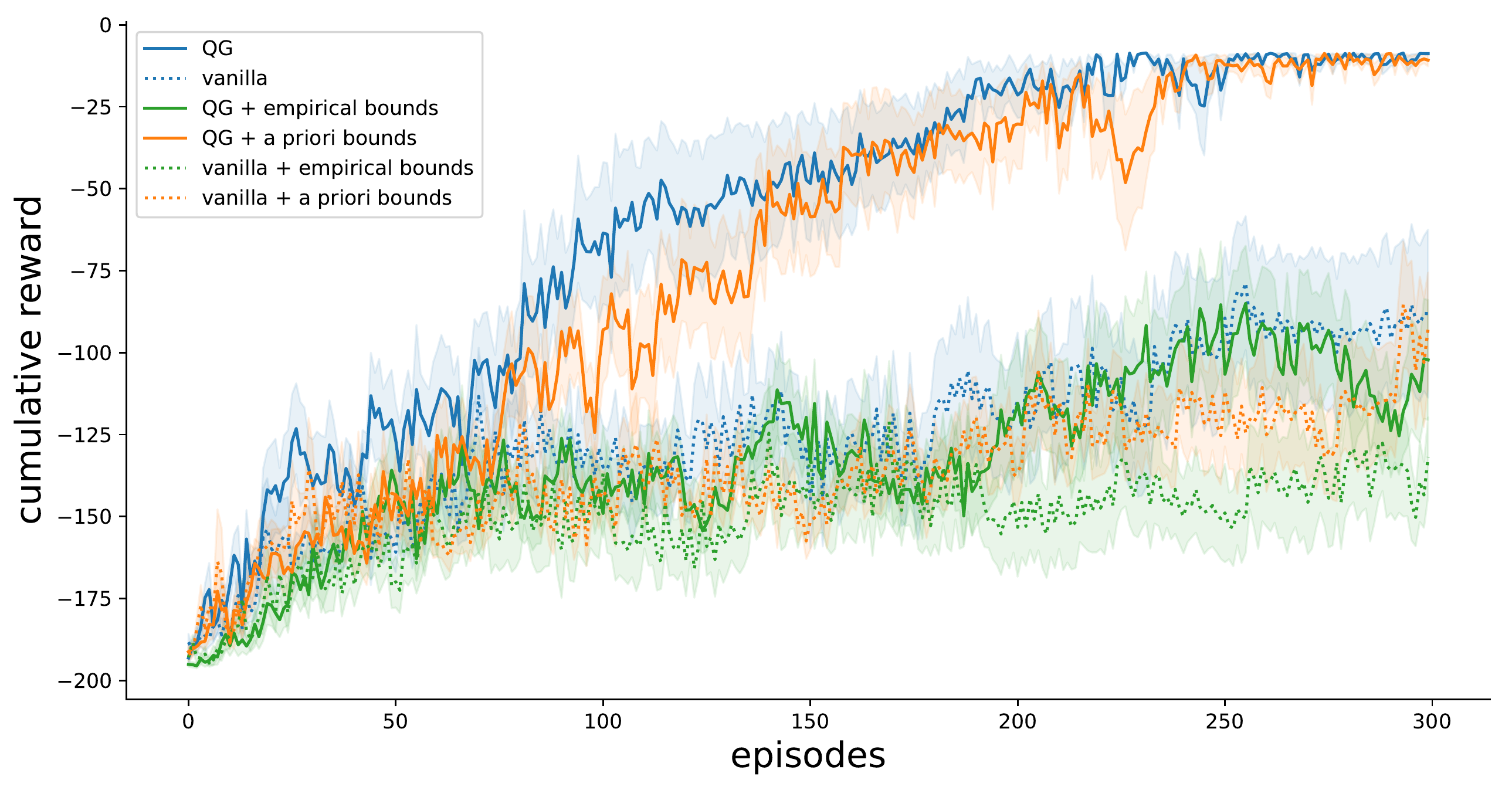}
	\end{center}
    \caption{Zero Actions (ZA, left) eliminate loose ends; trivial bounds are evaluated as baselines to our \qgraph-based bounds on the right.}
    \label{fig:baselines}
\end{figure}

We ran the following baselines to deepen our understanding of the previously reported effects:
In many settings a \textit{zero action} is known that does not change the agent's state (in our case it is the offset in position by zero meters).
Adding hypothetical transitions with the zero action after each physical transition ('vanilla-ZA') improves the structure of the data graph by turning loose ends into disconnected transitions.
Using zero actions in our method ('QG-ZA') not only improves the structure of the data graph but also spreads information in the form of lower bounds to predecessors in the \qgraph{}. \\
The results are shown in~\autoref{fig:baselines}.
Adding zero actions to vanilla DDPG does lead to an improvement, even without any \qgraph-bounded learning. 
This supports the importance of the data graph structure for Q-learning in general.
Also our method can be slightly improved by adding zero actions, but the largest performance gap is still between vanilla-ZA and our method. 
This indicates that while the data graph structure matters, the propagation of information through the \qgraph{} and the integration of lower bounds into TD-learning are the main causes for benefits from our method.

The next set of baselines was designed to evaluate how much influence the exact bounds have.
Bounded temporal difference learning could, besides our~\qgraph-based bounds, integrate two further types of lower and upper bounds:
\textit{A priori} bounds may be known in the case of a bounded reward function, see~\autoref{eq:minmaxQ}.
\textit{Empirical} bounds may seem like an alternative for correct a priori bounds: rather than using known bounds on the reward, these bounds could be estimated from experience. 
For the experiment, we used the lowest observed and highest observed rewards to compute bounds using~\autoref{eq:minmaxQ}.
Note that the true Q-values are guaranteed to lie within \qgraph-based bounds and correct a priori bounds, while empirical bounds might be too tight.
We combined~\ourmethod{} and vanilla DDPG with both types of bounds. When several bounds were available for one Q-value, the tightest upper and lower bound were chosen.
The results in~\autoref{fig:baselines} confirm that incorrect empirical bounds (green lines) have adverse effects on both methods, while a priori bounds do not seem to have any significant effect.
In particular, adding an upper a priori bound does not have a significant effect on our method.
We hypothesize that this may also be because the behavior of a Q-learning system differs for under- and over-estimated states: 
while under-estimated states may just never be visited (or rarely, depending on the type of exploration), over-estimated states are likely to be visited using the currently estimated optimal policy. 
Therefore, lower bounds correcting under-estimated states may be more important than upper bounds which would correct over-estimated states.
Overall, we conclude that the tight sample-specific lower bounds from our~\qgraph{} are key and much more informative than more general bounds.

\subsection{Limited Graph Capacity}
\label{subsec:exp-capacity}
In deep reinforcement learning, the replay memory is typically a FIFO-buffer ('first in, first out'), i.e.\ those elements that were added first are overwritten first when the buffer is full.
For a data graph, it is possible to delete single transitions but there are two possible effects:
On the one hand, some information from deleted transitions can be implicitly contained in its predecessors' Q-values on the \qgraph, which could imply that our method is more robust to small memory capacities.
On the other hand, cuts from deleted transitions can stop information propagation through the \qgraph, which could in turn slow down further progress.

We therefore empirically compared the drop in performance for vanilla DDPG and our \ourmethod{} with graph capacities of 1000 and 5000 transitions.
For comparison, the average unlimited graph contained roughly 30,000 unique transitions at the end of our 300 episode experiments.
As~\autoref{fig:capacity_and_nondeterminism} illustrates, a \qgraph-based method that is limited to only 1000 samples still performs on par with unlimited vanilla DDPG, while the vanilla DDPG performance decreases for a limit of 1000 transitions.

\subsection{Non-Deterministic Transitions}
\label{subsec:nondeterminism}
\begin{figure}
	\centering
	\includegraphics[width=0.49\linewidth]{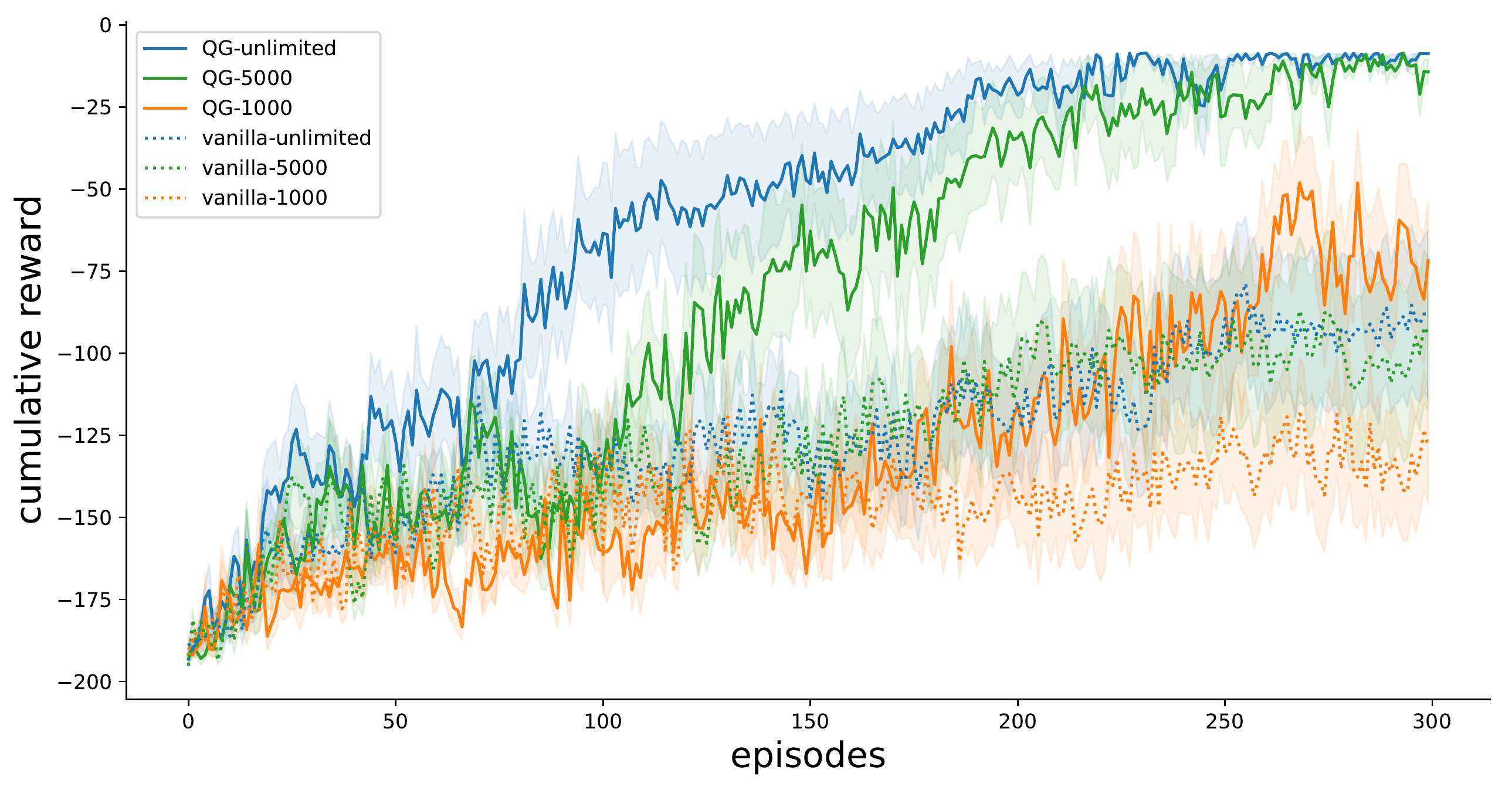}
	\includegraphics[width=0.49\linewidth]{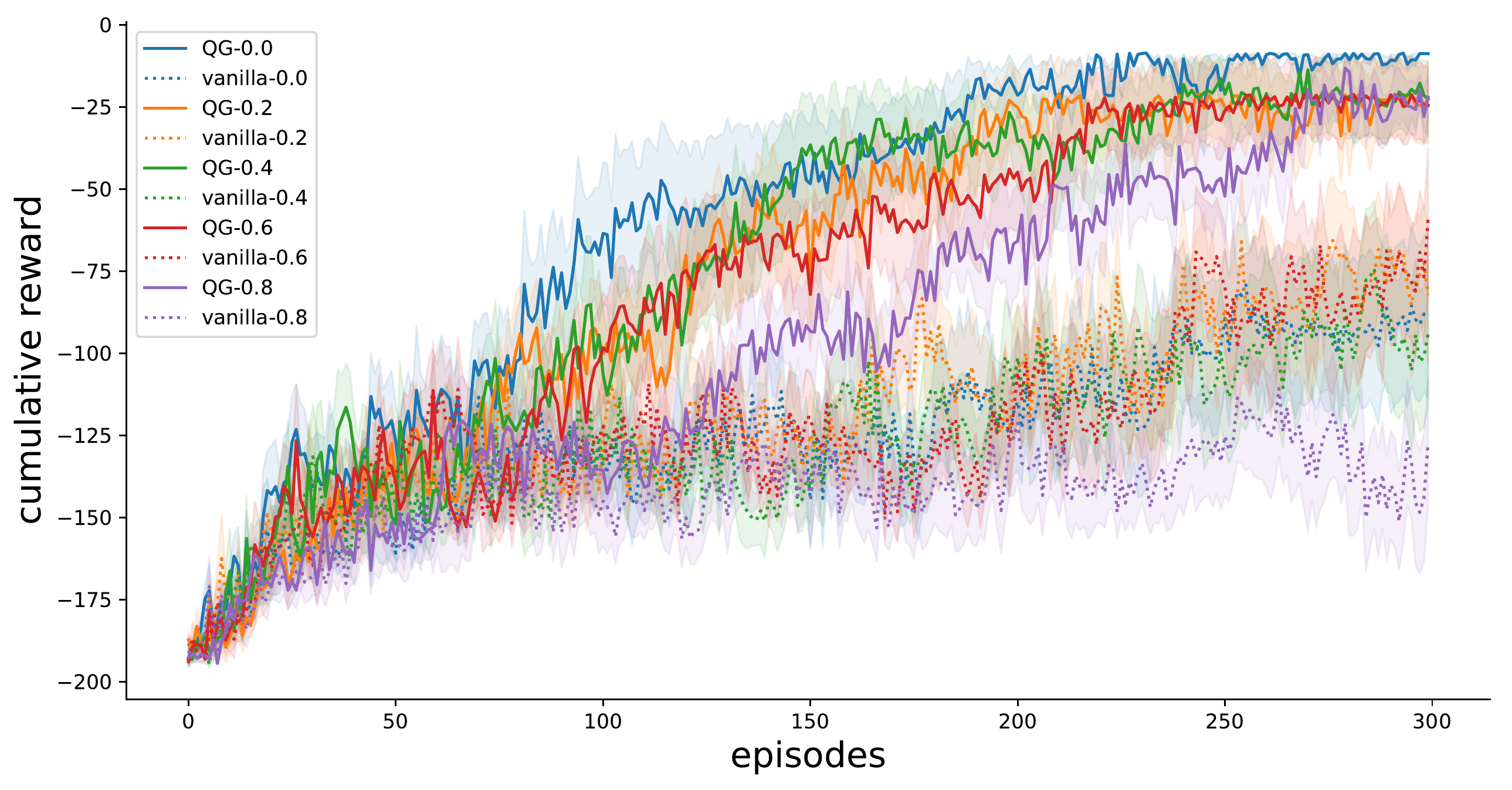}
	\caption{Performance with limited graph capacity (left) and increasingly non-deterministic transitions (right).}
	\label{fig:capacity_and_nondeterminism}
\end{figure}

As discussed in Section~\ref{sec:bounds}, the \qgraph-derived lower bounds are based on the assumption that all transitions are deterministic.
In case of non-deterministic transitions, correct lower bounds can be derived if for any state and any series of actions $\mathfrak{A}$, the empirical return $R$ that an agent can observe when following $\mathfrak{A}$ from $s$ differs by at most $\delta$.
In practice however, $\delta$ may not exist or be unknown.
We therefore empirically compare the results from~\autoref{subsec:exp-sampleEff-robustness} with increasing amounts of transition uncertainty.
To obtain the results shown in~\autoref{fig:capacity_and_nondeterminism}, each action was sampled from a Gaussian around the actor output with different $\sigma$: $\mathcal{N}(\pi(s), \sigma)$.
The results show that the performance generally drops with non-determinism for all methods, but the improvement of \ourmethod{} over vanilla DDPG remains significant.

\section{Conclusion}
From the observation that even for continuous state and action spaces, model-free off-policy deep reinforcement learning algorithms perform network updates on a finite set of transitions, we have developed a graph perspective on the replay memory that allows closer analysis.
Two types of data graph structures are clearly linked to soft divergence: 
non-terminal states without successors (\textit{loose ends}) and infinite loops with no path to a terminal state (\textit{disconnected states}).

Our method constructs a simplified MDP from a subgraph such that its exact Q-values can be computed by Q-iteration -- resulting in a \qgraph{}.
This subgraph does not contain loose ends, but we introduce so-called \textit{zero actions} which, if known, can be used to integrate loose ends into the \qgraph{} as well.\\
Q-values on the discrete simplified MDP associated with the \qgraph{} represent lower bounds for the Q-values in the original continuous MDP.
Enforcing these bounds in TD-learning empirically prevents cases of soft divergence on a continuous control task.

Preventing soft divergence as our method does, also increases sample efficiency on average and leads to the largest effect for unfavorable hyperparameters; i.e.\ our method increases robustness to adverse hyperparameters.
We have also demonstrated that the \qgraph{} can serve as an additional implicit memory holding information from transitions that have already been overwritten in the replay memory and thus, the algorithm is able to cope better with restricted memory capacity.
Empirically, the method also works in non-deterministic settings despite being derived under the assumption of deterministic transitions.

This work gives rise to a number of questions for future work:
(1) further bounds may exist, including data-driven or heuristic upper bounds; 
(2) the reward function most likely interacts with soft divergence and thus it may be possible to derive implications for reward shaping from our method; 
(3) exploration may benefit from current graph structure information;
(4) there may be further application-specific methods to integrate loose ends into the \qgraph{} structure, e.g. querying expert demonstrations.

\bibliographystyle{abbrvnat}
\bibliography{bib}

\newpage

\appendix
\section{PseudoCode}
\label{app:pseudocode}
\autoref{alg:train} describes the core of \ourmethod: one update step including the graph-based lower bounds, which were obtained from a \qgraph{} $\mathcal{G}$.
One possible implementation of a \qgraph{} which is iteratively constructed as new data comes in, is provided in~\autoref{alg:graph}.

\begin{algorithm}[H]
	\caption{\qgraph-bounded DDPG}
	\label{alg:train}
	\begin{algorithmic}[1]
	\Procedure{trainStep}{
	
	discount factor $\gamma$,
	
	actor network $\pi$, \Comment{mapping states to actions}
	
	critic network $\mathcal{Q}$, \Comment{predicting Q-values for state-action pairs}
	
	\qgraph{} $\mathcal{G}$, \Comment{see~\autoref{alg:graph}}
	
	a priori lower bound $\text{LB}^{AP}$, \Comment{A priori lower bound on Q-values if known. else $-\infty$}
	
	a priori upper bound $\text{UB}^{AP}$ 
	}\Comment{A priori upper bound on Q-values if known. else $+\infty$}	
	\item[]
	\State sample minibatch of $N$ transitions ${(s_i, a_i, s_i', r_i, t_i, \text{LB}_i^\mathcal{G})}_{i=0}^N$ from $\mathcal{G}$ \Comment{unknown lower bounds set to $-\infty$}
	\State $Q_\text{target}(s_i, a_i) = \begin{cases}
	r_i,& \text{if }s_i'\text{ is terminal (t)}\\
	r_i + \gamma \cdot \mathcal{Q}(s_i', \pi(s_i')),& \text{else}\\
	\end{cases}$ \Comment{classical Q targets, see~\autoref{eq:TD}}
	\State $\text{LB}_i = \text{max}(\text{LB}_i^\mathcal{G}, \text{LB}_i^{AP})$ \Comment{tightest available lower bound}
	\State $Q_\text{target}^B(s_i, a_i) = \text{min}(\text{UB}_i^{AP}, \text{max}(\text{LB}_i, Q_\text{target}(s_i, a_i)))$ \Comment{apply bounds, see~\autoref{eq:bounded_td}}
	\State $\mathcal{L}_C = \frac{1}{N} \sum_{i=0}^N (Q_\text{target}^B(s_i, a_i) - \mathcal{Q}(s_i, a_i))^2$ \Comment{DDPG Critic Loss, see~\autoref{eq:critic_loss}}
	\State $\mathcal{L}_A = - \frac{1}{N} \sum_{i=0}^{N}\mathcal{Q}(s_i,\pi(s_i))$ \Comment{DDPG Actor Loss}
	\State optimization step for both networks using $\mathcal{L}_A$ and $\mathcal{L}_C$
	\EndProcedure
\end{algorithmic}
\end{algorithm}

\newpage

\algrenewcommand\algorithmicensure{\textbf{Initialization:}}
\begin{algorithm}[H]
	\caption{Graph}
	\label{alg:graph}
	\begin{algorithmic}[1]
	\Ensure ~
	
	successors = \{\} \Comment{maps state $s$ to list of tuples ($s'$, $a$, $r$, $t$, LB$_Q$)}
	
	predecessors = \{\} \Comment{maps state $s'$ to list of tuples ($a$, $r$, $s$)}
	
	discount factor $\gamma$
	
	zero action ZA, if known
	
	capacity $\mathcal{C}$ \Comment{max. number of transitions to store}
	
	\item[]
	
	\Procedure{addTransition}{$s$, $a$, $s'$, $r$, $\mathfrak{t}$}
	\State add ($a$, $r$, $s$) to predecessors[$s'$] unless already exists
	\State LB$ = $\textsc{LBforNewTransition}($s$, $a$, $r$, $s'$, $\mathfrak{t}$) 
	\State add ($s'$, $a$, $r$, $\mathfrak{t}$, LB) to successors[$s$] unless already exists
	\If{LB $\neq$ \texttt{NaN}}
	\State \textsc{propagateLB($s$)} \Comment{Update predecessor bounds}
	\EndIf
	
	\If {capacity $\mathcal{C}$ reached}
	\State remove transition \Comment{e.g. first-in-first-out (FIFO)}
	\EndIf
	
	\If {Zero Action ZA known and $\mathfrak{t}=0$ and $s \neq s'$}
	\State \textsc{addTransition($s'$, ZA, $s'$, $\frac{r}{1-\gamma}$, $\mathfrak{t}=0$)}
	\EndIf
	
	\EndProcedure
	
	\item[]
	
	\Function{LBforNewTransition}{$s$, $a$, $r$, $s'$, $\mathfrak{t}$}
	\State $\text{LB} = \text{\texttt{NaN}}$ \Comment{lower bound unknown so far}
	
	\If{$\mathfrak{t}$}  \Comment{$s'$ is terminal state}
	\State $\text{LB} = \max(\text{LB}, r)$
	\EndIf
	
	\If{$s$ = $s'$}    \Comment{self-loop, e.g. zero action}
	\State $\text{LB} = \max(\text{LB}, \frac{r}{1 - \gamma})$
	\EndIf
	
	\If{larger loop with n transitions from $s$ detected}
	\State $\text{LB} = \max(\text{LB}, \frac{r_L}{1-\gamma^n})$ \Comment{see \autoref{eq:loops}}
	\EndIf
	
	\If{there are successor transitions from $s'$ with a lower bound}
	\State $\text{LB} = \max(\text{LB}, r + \gamma \cdot \max \{\text{lower bound LB' for transitions in successors[$s'$]} \})$
	\EndIf
	
	\State \Return LB \Comment{tightest lower bound}
	
	\EndFunction
	
	\item[]
	
	\Procedure{propagateLB}{start\_state}
	\State S = [start\_state] \Comment{list of states to visit}
	\While{states in S}
	\State s = S.pop(0) \Comment{remove and obtain first element in S}
	\If{s has predecessors and successors}
	\State LB$' = \max \{\text{lower bounds LB' for transitions in successors[$s$]}\}$ 
	\For {ps in predecessors[$s$]} \Comment{iterate predecessors of $s$}
	\State LB$_2 = r^\text{ps}_s + \gamma \cdot \text{LB}'$ \Comment{$r^\text{ps}_s$: reward for transition ps $\rightarrow$ s}
	\If {LB$_2 > $ existing bound for ps $\rightarrow$ s}
	\State update LB in transition ps $\rightarrow$ s
	\State S.add(ps)
	\EndIf
	\EndFor 
	\EndIf
	\EndWhile
	\EndProcedure
	\item[]
\end{algorithmic}
\end{algorithm}

\end{document}